\documentclass[5p,authoryear]{elsarticle} 
\usepackage{lineno}
\usepackage{hyperref}
\usepackage{amsmath}
\usepackage{amssymb}
\usepackage{verbatim}
\usepackage{multirow}
\usepackage{color,graphicx}
\usepackage{arydshln}

\usepackage{bm}
\usepackage{subfigure}

\providecommand{\Zxhreftb}[1]{Table~\ref{#1}}
\providecommand{\zxhreftb}[1]{Table~\ref{#1}}
\providecommand{\zxhreffig}[1]{Fig.~\ref{#1}}

\providecommand{\zxhcolor}[2]{{\color{red}#2}}

\modulolinenumbers[200]

\journal{Medical Image Analysis}

\begin{document}

\begin{frontmatter}
\title{Evaluation of Algorithms for Multi-Modality Whole Heart Segmentation: \\An Open-Access Grand Challenge}

\author{Xiahai~Zhuang$^{1,2,}$*} \ead[url]{zxh@fudan.edu.cn}
\author{Lei~Li$^{3,}$*} \ead[url]{lilei.sky@sjtu.edu.cn}
\author{Christian~Payer$^{4}$} 
\author{Darko \v{S}tern$^{5}$} 
\author{Martin~Urschler$^{5}$} 
\author{Mattias P. Heinrich$^{6}$} 
\author{Julien~Oster$^{7}$} 
\author{Chunliang Wang$^{8}$} 
\author{\"{O}rjan Smedby$^{8}$} 
\author{Cheng Bian$^{9}$} 
\author{Xin Yang$^{10}$} 
\author{Pheng-Ann Heng$^{10}$} 
\author{Aliasghar Mortazi$^{11}$} 
\author{Ulas Bagci$^{11}$} 
\author{Guanyu Yang$^{12}$} 
\author{Chenchen Sun$^{12}$} 
\author{Gaetan Galisot$^{13}$} 
\author{Jean-Yves Ramel$^{13}$} 
\author{Thierry Brouard$^{13}$} 
\author{Qianqian~Tong$^{14}$} 
\author{Weixin Si$^{15}$} 
\author{Xiangyun Liao$^{16}$} 
\author{Guodong Zeng$^{17}$} 
\author{Zenglin Shi$^{17}$} 
\author{Guoyan Zheng$^{17}$} 
\author{Chengjia~Wang$^{18,19}$} 
\author{Tom MacGillivray$^{19}$} 
\author{David Newby$^{18,19}$}
\author{Kawal Rhode$^{20}$} 
\author{Sebastien Ourselin$^{20}$}
\author{Raad Mohiaddin$^{21,22}$}
\author{Jennifer Keegan$^{21,22}$}
\author{David Firmin$^{21,22}$}
\author{Guang~Yang$^{21,22,}$*} \ead[url]{g.yang@imperial.ac.uk}

\address{$^1$School of Data Science, Fudan University, 200433, Shanghai, China\\[0.5ex]
$^2$Fudan-Xinzailing Joint Research Center for Big Data, Fudan University, 200433, Shanghai, China \\[0.5ex]
$^3$School of Biomedical Engineering, Shanghai Jiao Tong University, 200240, Shanghai, China\\[0.5ex]
$^4$Institute of Computer Graphics and Vision, Graz University of Technology, 8010, Graz, Austria \\[0.5ex]
$^5$Ludwig Boltzmann Institute for Clinical Forensic Imaging, 8010, Graz, Austria \\[0.5ex]
$^6$Institute of Medical Informatics, University of Lubeck, 23562, Lubeck, Germany \\[0.5ex]
$^7$Inserm, Universit\'{e} de Lorraine, U1254, IADI, Nancy, France\\[0.5ex]
$^8$School for Technology and Health, KTH Royal Institute of Technology, SE-10044, Stockholm, Sweden\\[0.5ex]
$^9$School of Biomed. Eng., Health Science Centre, Shenzhen University, 518060, Shenzhen, China\\[0.5ex]
$^{10}$Dept. of Comp. Sci. and Eng., The Chinese University of Hong Kong, Hong Kong, China\\[0.5ex]
$^{11}$Center for Research in Computer Vision (CRCV), University of Central Florida, 32816, Orlando, U.S. \\[0.5ex]
$^{12}$School of Computer Science and Engineering, Southeast University, 210096, Nanjing, China\\[0.5ex]
$^{13}$LIFAT (EA6300), Universit\'{e} de Tours, 64 avenue Jean Portalis, 37200, Tours, France\\[0.5ex]
$^{14}$School of Computer Science, Wuhan University, 430072, Wuhan, China\\[0.5ex]
$^{15}$Guangdong Provincial Key Laboratory of Computer Vision and Virtual Reality Technology, SIAT, Shenzhen, China\\[0.5ex]
$^{16}$Shenzhen Institutes of Advanced Technology, Chinese Academy of Sciences, 518055, Shenzhen, China\\[0.5ex]
$^{17}$Institute for Surgical Technology \& Biomechanics, University of Bern, 3014, Bern, Switzerland\\[0.5ex]
$^{18}$BHF Centre for Cardiovascular Science, University of Edinburgh, Edinburgh, U.K.\\[0.5ex]
$^{19}$Edinburgh Imaging Facility QMRI, University of Edinburgh, Edinburgh, U.K.\\[0.5ex]
$^{20}$Department of Imaging Sciences \& Biomedical Engineering, Kings College London, London, U.K.\\[0.5ex]
$^{21}$Cardiovascular Research Centre, Royal Brompton Hospital, SW3 6NP, London, U.K.\\[0.5ex]
$^{22}$National Heart and Lung Institute, Imperial College London, London, SW7 2AZ, London, U.K.}

\begin{abstract}
Knowledge of whole heart anatomy is a prerequisite for many clinical applications.
Whole heart segmentation (WHS), which delineates substructures of the heart, can be very valuable for modeling and analysis of the anatomy and functions of the heart.
However, automating this segmentation can be arduous due to the large variation of the heart shape, and different image qualities of the clinical data.
To achieve this goal, a set of training data is generally needed for constructing priors or for training.
In addition, it is difficult to perform comparisons between different methods, largely due to differences in the datasets and evaluation metrics used.
This manuscript presents the methodologies and evaluation results for the WHS algorithms selected from the submissions to the Multi-Modality Whole Heart Segmentation (MM-WHS) challenge, in conjunction with MICCAI 2017.
The challenge provides 120 three-dimensional cardiac images covering the whole heart, including 60 CT and 60 MRI volumes, all acquired in clinical environments with manual delineation.
Ten algorithms for CT data and eleven algorithms for MRI data, submitted from twelve groups, have been evaluated.
The results show that many of the deep learning (DL) based methods achieved high accuracy, even though the number of  training datasets were limited.
A number of them also reported poor results in the blinded evaluation, probably due to overfitting in their training.
The conventional algorithms, mainly based on multi-atlas segmentation, demonstrated robust and stable performance, even though the accuracy is not as good as the best DL method in CT segmentation.
The challenge, including provision of the annotated training data and the blinded evaluation for submitted algorithms on the test data, continues as an ongoing benchmarking resource via its homepage (\url{www.sdspeople.fudan.edu.cn/zhuangxiahai/0/mmwhs/}).
\end{abstract}


\end{frontmatter}
\linenumbers

\section{Introduction}

According to the World Health Organization, cardiovascular diseases (CVDs) are the leading cause of death globally \citep{book/who/Mendis2011}.
Medical imaging has revolutionized the modern medicine and healthcare, and the imaging and computing technologies become increasingly important for the diagnosis and treatments of CVDs.
Computed tomography (CT), magnetic resonance imaging (MRI), positron emission tomography (PET), single photon emission computed tomography (SPECT), and ultrasound (US) have been used extensively for physiologic understanding and diagnostic purposes in cardiology \citep{journal/jei/Kang2012}.
Among these, CT and MRI are particularly used to provide clear anatomical information of the heart.
Cardiac MRI has the advantages of being free from ionizing radiation, acquiring images with great contrast between soft tissues and relatively high spatial resolutions \citep{journal/ii/Nikolaou2011}.
On the other hand, cardiac CT is fast, low cost, and generally of high quality \citep{journal/Heart/Roberts2008}.

\begin{figure*}[t]\center
   \subfigure[] {\includegraphics[width=0.49\textwidth]{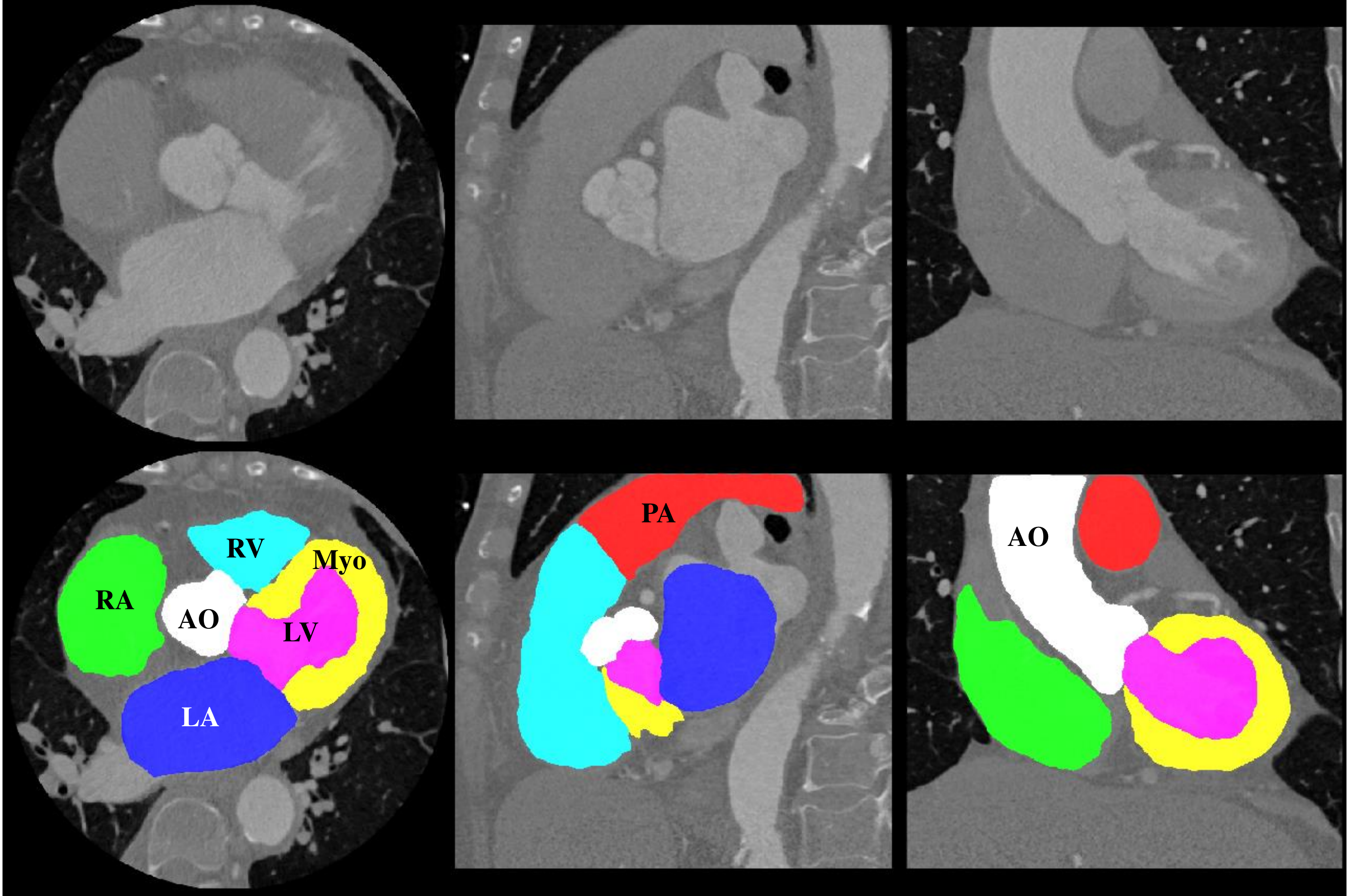}}
   \subfigure[] {\includegraphics[width=0.49\textwidth]{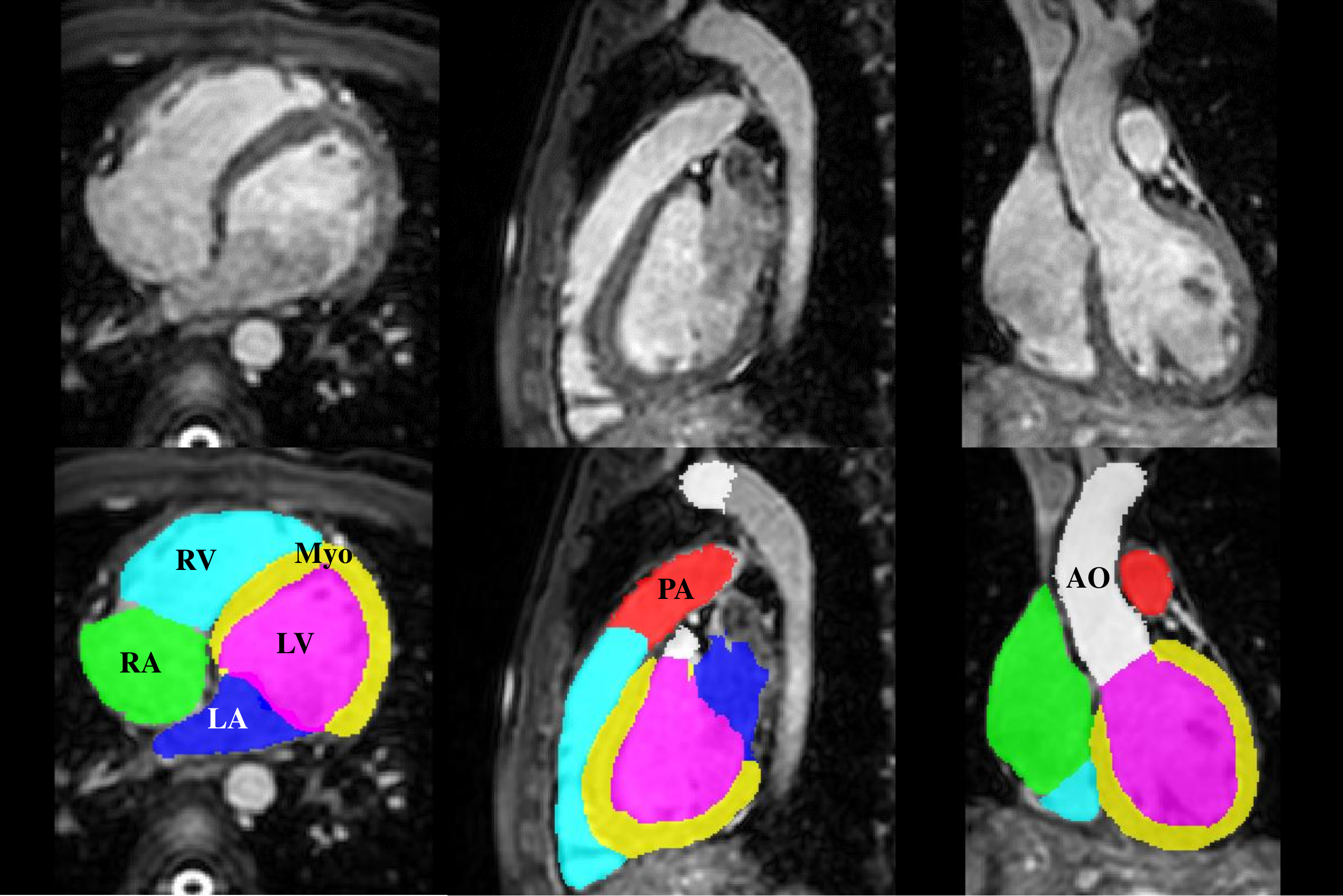}}
   \caption{Examples of cardiac images and WHS results: (a) displays the three orthogonal views of a cardiac CT image and its corresponding WHS result, (b) is from a cardiac MRI image and its WHS.
   LV: left ventricle; RV: right ventricle; LA: left atrium; RA: right atrium; Myo: myocardium of LV; AO: ascending aorta; PA: pulmonary artery.
   }
\label{fig:examplemrict}\end{figure*}

To quantify the morphological and pathological changes, it is commonly a prerequisite to segment the important structures from the cardiac medical images.
Whole heart segmentation (WHS) aims to extract each of the individual whole heart substructures, including
the left ventricle (LV), right ventricle (RV), left atrium (LA), right atrium (RA), myocardium of LV (Myo), ascending aorta (AO) or the whole aorta, and the pulmonary artery (PA) \citep{journal/jhe/Zhuang13},
 as \zxhreffig{fig:examplemrict} shows. 
The applications of WHS are ample.
The results can be used to directly compute the functional indices such as ejection fraction.
Additionally, the geometrical information is useful in surgical guidance such as in radio-frequency ablation of the LA.
However, the manual delineation of whole heart is labor-intensive and tedious, needing almost 8 hours for a single subject \citep{journal/mia/Zhuang2016}.
Thus, automating the segmentation from multi-modality images, referred to as MM-WHS, is highly desired but still challenging, mainly due to the following reasons \citep{journal/jhe/Zhuang13}.
First, the shape of the heart varies largely in different subjects or even for the same subject at different cardiac phases, especially for those with pathological and physiological changes.
Second, the appearance and image quality can be variable.
For example, the enhancement patterns of the CT images can vary significantly for different scanners or acquisition sessions.
Also, motion artifacts, poor contrast-to-noise ratio and signal-to-noise ratio, commonly presented in the clinical data, can significantly deteriorate the image quality and consequently challenge the task.

\subsection{State-of-the-art for Whole Heart Segmentation}\label{stateofwhs}
In the last ten years, a variety of WHS techniques have been proposed for cardiac CT and MRI data.
The detailed reviews of previously published algorithms can be found in \citet{journal/jei/Kang2012}, \citet{journal/jhe/Zhuang13} and \citet{journal/mrmpbm/Peng2016}.
\citet{journal/jei/Kang2012} reviewed several modalities and corresponding segmentation algorithms for the diagnosis and treatments of CVDs.
They summarized the roles and characteristics of different modalities of cardiac imaging and the parameter correlation between them.
In addition, they categorized the WHS techniques into four kinds, i.e., (1) boundary-driven techniques, (2) region-based techniques, (3) graph-cuts techniques, and (4) model fitting techniques.
The advantages and disadvantages of each category were analyzed and summarized.
\citet{journal/jhe/Zhuang13} discussed the challenges and methodologies of the fully automatic WHS.
Particularly, the work summarized two key techniques, i.e., the construction of prior models and the fitting procedure for segmentation propagation, for achieving this goal.
Based on the types of prior models, the segmentation methods can be divided into two groups, namely the deformable model based methods and the atlas-based approaches;
 and the fitting procedure can be decomposed into three stages, including localizing the whole heart, initializing the substructures, and refining the boundary delineation.
Thus, this review paper mainly analyzes the algorithms based on the classification of prior models and fitting algorithms for the WHS from different modality images.
\citet{journal/mrmpbm/Peng2016} reviewed both the methodologies of WHS and the structural and functional indices of the heart for clinical assessments.
In their work, the WHS approaches were classified into three categories, i.e., image-driven techniques, model-driven techniques, and direct estimation.

The three topic review papers mentioned above mainly cover the publications before 2015. A collection of recent works not included by them are summarized in \Zxhreftb{tb:table:review}.
Among these works,
\citep{journal/mp/zhuang2015} proposed an atlas ranking and selection scheme based on conditional entropy for the multi-atlas based WHS of CT.
\citet{journal/ro/Zhou2017} developed a set of CT atlases labeled with 15 cardiac substructures.
These atlases were then used for automatic WHS of CT via the multi-atlas segmentation (MAS) framework.
\citet{journal/Neurocomputing/Cai2017} developed a method with window width-level adjustment to pre-process CT data, which generates images with clear anatomical structures for WHS.
They applied a Gaussian filter-based multi-resolution scheme to eliminate the discontinuity in the down-sampling decomposition for whole heart image registration.
\citet{conf/icfimh/Zuluaga2013} developed a MAS scheme for both CT and MRI WHS.
The proposed method ranked and selected optimal atlases based on locally normalised cross correlation.
\citet{conf/sip/Pace2015} proposed a patch-based interactive algorithm to extract the heart based on a manual initialization from experts.
The method employs active learning to identify the areas that require user interaction.
\citet{journal/mia/Zhuang2016} developed a multi-modality MAS framework for WHS of cardiac MRI, which used a set of atlases built from both CT and MRI.
The authors proposed modality invariant metrics for computing the global image similarity and the local similarity.
The global image similarity was used to rank and select atlases, from the multi-modality atlas pool, for segmenting a target image;
and the local similarity metrics were proposed for the patch-based label fusion, where a multi-scale patch strategy was developed to obtain a promising performance.

In conclusion, WHS based on the MAS framework, referred to as MA-WHS, has been well researched in recent years.
MAS segments an unknown target image by propagating and fusing the labels from multiple annotated atlases using registration.
The performance relies on the registration algorithms for label propagation and the fusion strategy to combine the segmentation results from the multiple atlases.
Both these two key steps are generally computationally expensive.

Recently, a number of deep learning (DL)-based methods have shown great promise in medical image analysis.
They have obtained superior performance in various imaging modalities and different clinical applications \citep{conf/miccai/Roth2014,journal/arbe/Shen2017}.
For cardiac segmentation, \citet{journal/mia/Avendi2016} proposed a DL algorithm for LV segmentation.
\citet{journal/mia/Ngo2017} trained multiple layers of deep belief network to localize the LV, and to define the endocardial and epicardial borders, followed by the distance regularised level set.
Recently, \citet{journal/jmri/Tan2018} designed a fully automated convolutional neural network (CNN) architecture for pixel-wise labeling of both the LV and RV with impressive performance.
DL methods have potential of providing faster and more accurate segmentation, compared to the conventional approaches, such as the deformable model based segmentation and MAS method.
However, little work has been reported to date using DL for WHS, probability due to the limitation of training data and complexity of the segmentation task.

\zxhreftb{tb:table:challenge} summarizes the recent open access datasets for cardiac segmentation, which mainly focus on specific substructures of the heart.
\citet{journal/ij/radau2008,conf/stacom/Suinesiaputra2011,journal/MedAI/Petitjean2015,journal/tmi/Bernard2018} 
organized the challenges for segmenting the left, right or full ventricles.
\citet{link/HVSMR2016} organized a challenge for the segmentation of blood pool and myocardium from 3D MRI data.
This work aims to offer pre-procedural planning of children with complex congenital heart disease.
\citet{journal/tmi/Tobon2015,journal/MedAI/karim2018} and \citet{link/LAseg2018}
provided data for benchmarking algorithms of LA or LA wall segmentation for patients suffering from atrial fibrillation.

\begin{table*} [t] \center
    \caption{
    Summary of previous WHS methods for multi-modality images. Here, the abbreviations are as follows,
    PIS: patch-based interactive segmentation;
    FIMH: International Conference on Functional Imaging and Modeling of the Heart;
    MICCAI: International Conference on Medical Image Computing and Computer-assisted Intervention;
    MedPhys: Medical Physics;
    MedIA: Medical Image Analysis;
    RadiotherOncol: Radiotherapy and Oncology.}
\label{tb:table:review}
{\small
\begin{tabular}{ l  l lll }\hline
Reference       & Data &  Method  & Runtime & Dice \\
\hline
\citet{conf/icfimh/Zuluaga2013}, FIMH        & 8 CT, 23 MRI     & MAS                       & 60 min, 30 min       & $0.89 \pm 0.04$, $0.91 \pm 0.03$\\
\citet{journal/mp/zhuang2015}, MedPhys       & 30 CT            & MAS                       & 13.2 min             & $0.92 \pm 0.02$ \\
\citet{conf/sip/Pace2015}, MICCAI            & 4 MRI            & PIS + Active learning     & 60 min               & N/A           \\
\citet{journal/mia/Zhuang2016}, MedIA        & 20 CT + 20 MRI   & Multi-modality MAS        & 12.58 min            & $0.90 \pm 0.03$ \\
\citet{journal/ro/Zhou2017}, RadiotherOncol  & 31 CT            & MAS                       & 10 min               & $0.77 \pm 0.07$ \\
\citet{journal/Neurocomputing/Cai2017}, Neurocomputing   & 14 CT            & Gaussian filter-based     & N/A                  & $0.77 \pm 0.07$ \\
\hline
\end{tabular} }
\end{table*}

\begin{table*} [htp] \center
    \caption{
    Summary of the previous challenges related to cardiac segmentation from MICCAI society.
     }
\label{tb:table:challenge}
{\small
\begin{tabular}{ lllll} \hline
Organizers/refernece       &  Year & Data & Target & Pathology \\
\hline
\citet{journal/ij/radau2008}          & 2009 & 45 cine MRI          & LV                    & hypertrophy, infarction\\
\citet{conf/stacom/Suinesiaputra2011} & 2011 & 200 cine MRI         & LV                    & myocardial infarction \\
\citet{journal/MedAI/Petitjean2015}   & 2012 & 48 cine MRI          & RV                    & congenital heart disease \\
\citet{journal/tmi/Tobon2015}         & 2013 & 30 CT + 30 MRI       & LA                    & atrial fibrillation\\
\citet{journal/MedAI/karim2018}       & 2016 & 10 CT + 10 MRI       & LA wall               & atrial fibrillation\\
\citet{link/HVSMR2016}                & 2016 & 20 MRI               & Blood pool, Myo       & congenital heart disease \\
\citet{journal/tmi/Bernard2018}       & 2017 & 150 cine MRI         & Ventricles            & infarction, dilated/ hypertrophic \\[-0.2ex]
  & & & &cardiomyopathy, abnormal RV\\
\citet{link/LAseg2018}                & 2018 & 150 LGE-MRI          & LA                    & atrial fibrillation \\
\hline
\end{tabular} }\\
\end{table*}

\subsection{Motivation and Contribution}\label{motivation}

Due to the above mentioned challenges, we organized the competition of MM-WHS, providing 120 multi-modality whole heart images for developing new WHS algorithms, as well as validating existing ones.
We also presented a fair evaluation and comparison framework for participants.
In total, twelve groups who submitted their results and methods were selected, and they all agreed to contribute to this work,
a benchmark for WHS of two modalities, i.e., CT and MRI.
In this work, we introduce the related information, elaborate on the methodologies of these selective submissions, discuss the results and provide insights to the future research.


The rest of this paper is organised as follows. Section \ref{material} provides details of the materials and evaluation framework. Section \ref{methods} introduces the evaluated methods for benchmarking.
Section \ref{result} presents the results, followed by discussions in Section \ref{discussion}.
We conclude this work in Section \ref{conclusion}.

\begin{table*} [t] \center
    \caption{Summary of submitted methods.
    Asterisk (*) indicates the results that were submitted after the challenge deadline.
    }
\label{tb:table:TeamName}
{\footnotesize
\begin{tabular}{| l l p{5.3cm} | l l p{5.3cm} |}\hline
Teams &  Tasks &  Key elements in methods &Teams &  Tasks &  Key elements in methods \\\hline
GUT     & CT, MRI    & Two-step CNN, combined with anatomical label configurations. &
UOL     & MRI        & MAS and discrete registration, to adapt the large shape variations.\\
KTH     & CT, MRI    & Multi-view U-Nets combining hierarchical shape prior.&
CUHK1   & CT, MRI    & 3D FCN with the gradient flow optimization and Dice loss function. \\
SEU     & CT         & Conventional MAS-based method.&
CUHK2   & CT, MRI    & Hybrid loss guided FCN.\\
UCF     & CT, MRI    & Multi-object multi-planar CNN with an adaptive fusion method.&
UT      & CT, MRI    & Local probabilistic atlases coupled with a topological graph.\\
SIAT    & CT, MRI    & 3D U-Net learning learn multi-modality features.&
UB2$^*$ & MRI        & Multi-scale fully convolutional Dense-Nets.\\
UB1$^*$ & CT, MRI    & Dilated residual networks.&
UOE$^*$ & CT, MRI    & Two-stage concatenated U-Net.\\
\hline
\end{tabular} }
\end{table*}

\section{Materials and setup} \label{material}

\subsection{Data Acquisition}

The cardiac CT/CTA data were acquired from two state-of-the-art 64-slice CT scanners (Philips Medical Systems, Netherlands) using a standard coronary CT angiography protocol at two sites in Shanghai, China.
All the data cover the whole heart from the upper abdomen to the aortic arch.
The in-plane resolution of the axial slices is $0.78\!\times\!0.78$  mm, and the average slice thickness is 1.60 mm.

The cardiac MRI data were obtained from two hospitals in London, UK.
One set of data were acquired from St. Thomas Hospital on a 1.5T Philips scanner (Philips Healthcare, Best, The Netherlands),
and the other were from Royal Brompton Hospital on a Siemens Magnetom Avanto 1.5T scanner (Siemens Medical Systems, Erlangen, Germany).
In both sites we used the 3D balanced steady state free precession (b-SSFP) sequence for whole heart imaging,
and realized free breathing scans by enabling a navigator beam before data acquisition for each cardiac phase.
The data were acquired at a resolution of around $(1.6\!\!\sim\!\!2)\!\times\!(1.6\!\!\sim\!\!2)\!\times\!(2\!\!\sim\!\!3.2)$ mm, and reconstructed to half of its acquisition resolution, i.e., about $(0.8\!\!\sim\!\!1)\!\times\!(0.8\!\!\sim\!\!1)\!\times\!(1\!\!\sim\!\!1.6)$ mm.

Both cardiac CT and cardiac MRI data were acquired in real clinical environment.
The pathologies of patients cover a wide range of cardiac diseases, including myocardium infarction, atrial fibrillation, tricuspid regurgitation, aortic valve stenosis, Alagille syndrome, Williams syndrome, dilated cardiomyopathy, aortic coarctation, Tetralogy of Fallot.
The subjects for MRI scans also include a small number of normal controls.

All the CT and MRI data have been anonymized in agreement with the local regional ethics committee before being released to the MM-WHS challenge.
In total, we provided 120 multi-modality whole heart images from multiple sites, including 60 cardiac CT and 60 cardiac MRI.
Note that the data were collected from clinical environments, so the image quality was variable.
This enables to assess the validation and robustness of the developed algorithms with representative clinical data, rather than with selected best quality images.

\subsection{Definition and Gold Standard}
The WHS studied in this work aims to delineate and extract the seven substructures of the heart, into separate individuals \citep{journal/jhe/Zhuang13}.
These seven structures include the following,
\begin{enumerate}
 \item[(1)] the LV blood cavity, also referred to as LV;
 \item[(2)] the RV blood cavity, also referred to as RV;
 \item[(3)] the LA blood cavity, also referred to as LA;
 \item[(4)] the RA blood cavity, also referred to as RA;
 \item[(5)] the myocardium of the LV (Myo) and the epicardium (Epi), defined as the epicardial surface of the LV;
 \item[(6)] the AO trunk from the aortic valve to the superior level of the atria, also referred to as AO;
 \item[(7)] the PA trunk from the pulmonary valve to the bifurcation point, also referred to as PA.
\end{enumerate}
The four blood pool cavities, i.e., LV, RV, LA and RA, are also referred to as the four chambers.

Manual labeling was adopted for generating the gold standard segmentation.
They were done by clinicians or by students majoring in biomedical engineering or medical physicists who were familiar with the whole heart anatomy,
slice-by-slice using the ITK-SNAP software \citep{journal/Neuroimage/Yushkevich2006}.
Each manual segmentation result was examined by a senior researchers specialized in cardiac imaging with experience of more than five years, and modifications have been take if revision was necessary.
Also, the sagittal and coronal views were visualised simultaneously to check the consistency and smoothness of the segmentation,
although the manual delineation was mainly performed in the axial views.
For each image, it takes approximately 6 to 10 hours for the observer to complete the manual segmentation of the whole heart.

\subsection{Evaluation Metrics}

We employed four widely used metrics to evaluate the accuracy of a segmentation result, including the
Dice score \citep{journal/pami/Kittler98}, Jaccard index \citep{journal/Jaccard1901}, surface-to-surface distance (SD), and Hausdorff  Distance (HD).
For WHS evaluation, we adopted the generalized version of them, the normalized metrics with respect to the size of substructures.
They are expected to provide more objective measurements \citep{journal/tmi/CrumCH06,journal/jhe/Zhuang13}.

For each modality, the data were split into two sets, i.e., the training set (20 CT and 20 MRI) and the test set (40 CT and 40 MRI).
For the training data, both the images and the corresponding gold standard were released to the participants for building, training and cross-validating their models.
For the test data, only the CT and MRI images were released.
Once the participants developed their algorithms, they could submit their segmentation results on the test data to the challenge moderators for a final independent evaluation.
To make a fair comparison, the challenge organizers only allowed maximum of two evaluations for one algorithm.

\subsection{Participants}
Twelve algorithms (teams) were selected for this benchmark work.
Nine of them provided results for both CT and MRI data, one experimented only on the CT data and two worked solely on the MRI data.

All of the 12 teams agreed to include their results in this paper.
To simplify the description below, we used the team abbreviations referring to both the teams and their corresponding methods and results.
The evaluated methods are elaborated on in Section \ref{methods},
 and the key contributions of the teams are summarized in \zxhreftb{tb:table:TeamName}.
Note that the three methods, indicated with Asterisk (*), were submitted after the challenge deadline for performance ranking.

\section{Evaluated Methods} \label{methods}
In this section, we elaborate on the twelve benchmarked algorithms. \zxhreftb{tb:table:TeamName} provides the summary for reference.

\subsection{Graz University of Technology (GUT)}

\citet{conf/stacom/Payer2017} proposed a fully automatic whole heart segmentation, based on multi-label CNN and using volumetric kernels, which consists of two separate CNNs:
one to localize the heart, referred to as localization CNN, and the other to segment the fine detail of the whole heart structure within a small region of interest (ROI),
referred to as segmentation CNN.
The localization CNN is designed to predict the approximate centre of the bounding box around all heart substructures, based on the U-Net \citep{conf/miccai/Ronneberger2015} and heatmap regression \citep{conf/miccai/Payer2016}.
A fixed physical size ROI is then cropped around the predicted center, ensuring that it can enclose all interested substructures of the heart.
Within the cropped ROI, the multi-label segmentation CNN predicts the label of each pixel.
In this method, the  segmentation CNN works on high-resolution ROI, while the localization CNN works on the low resolution images.
This two-step CNN pipeline helps to mitigate the intensive memory and runtime generally required by the volumetric kernels equipped 3D CNNs.



\subsection{University of Lubeck (UOL)}

\begin{figure}[t]\center
 \includegraphics[width=0.57\textwidth]{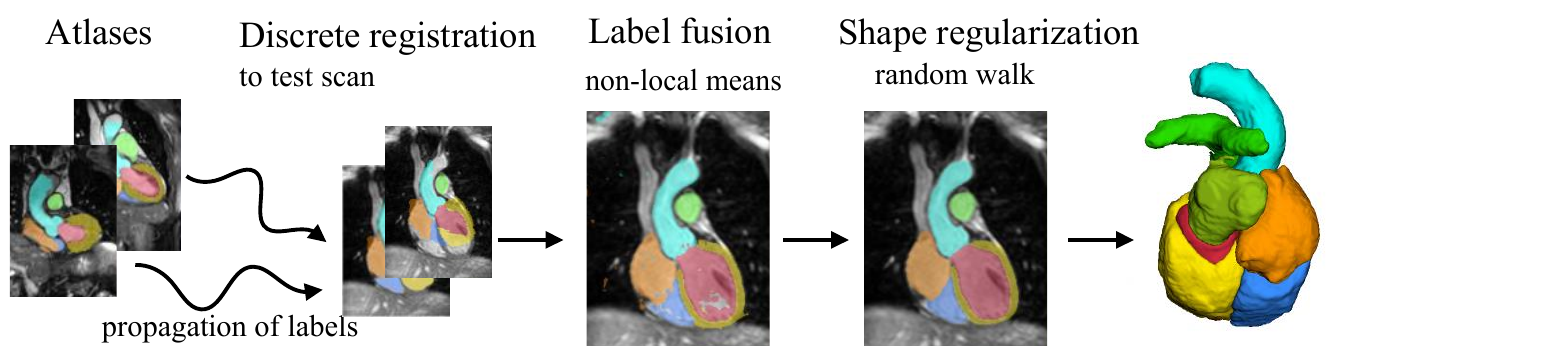}
   \caption{Multi-atlas registration and label fusion with regularization proposed by \citet{conf/stacom/Heinrich2017}.}
\label{fig:method:UOL}\end{figure}

\citet{conf/stacom/Heinrich2017} proposed a multi-atlas registration approach for WHS of MRI, as \zxhreffig{fig:method:UOL} shows.
This method adopts a discrete registration, which can capture large shape variations across different scans \citep{journal/tmi/Heinrich2013}.
Moreover, it can ensure the alignment of anatomical structures by using dense displacement sampling
 and graphical model-based optimization \citep{conf/miccai/Heinrich2013}.
Due to the use of contrast-invariant features \citep{journal/tbe/Xu2016}, the multi-atlas registration can implicitly deal with the challenging varying intensity distributions due to different acquisition protocols.
Within this method, one can register all the training atlases to an unseen test image.
The warped atlas label images are then combined by means of weighted label fusion.
Finally, an edge-preserving smoothing of the generated probability maps is performed using the multi-label random walk algorithm, as implemented and parameterized in \citet{conf/miccai/Heinrich2016}.

\begin{table*} [t] \center
    \caption{
    Results of the ten evaluated algorithms on CT dataset.
    SD: surface-to-surface distance;
    HD: Hausdorff  Distance;
    DL: deep learning-based method;
    MAS: conventional method based on multi-atlas segmentation.
    Asterisk (*) indicates the results were submitted after the challenge deadline.
    }
\label{tb:table:result_ct}
{\footnotesize
\begin{tabular}{ l|lllll}\hline
Teams &  Dice &  Jaccard &   SD (mm) &   HD (mm)   &  DL/MAS\\
\hline
GUT      &  \bm{$ 0.908 \pm 0.086 $}  & \bm{$ 0.832 \pm 0.037 $}  &\bm{$ 1.117 \pm 0.250 $}& \bm{$ 25.242 \pm 10.813 $}& DL \\
KTH      &  $ 0.894 \pm 0.030 $  &    $ 0.810 \pm 0.048 $&  $ 1.387 \pm 0.516 $&    $ 31.146 \pm 13.203 $&  DL \\
CUHK1    &  $ 0.890 \pm 0.049 $  &    $ 0.805 \pm 0.074 $&  $ 1.432 \pm 0.590 $&    $ 29.006 \pm 15.804 $&  DL \\
CUHK2    &  $ 0.886 \pm 0.047 $  &    $ 0.798 \pm 0.072 $&  $ 1.681 \pm 0.593 $&    $ 41.974 \pm 16.287 $&  DL \\
UCF      &  $ 0.879 \pm 0.079 $  &    $ 0.792 \pm 0.106 $&  $ 1.538 \pm 1.006 $&    $ 28.481 \pm 11.434 $&  DL \\
SEU      &  $ 0.879 \pm 0.023 $  &    $ 0.784 \pm 0.036 $&  $ 1.705 \pm 0.399 $&    $ 34.129 \pm 12.528 $&  MAS \\
SIAT     &  $ 0.849 \pm 0.061 $  &    $ 0.742 \pm 0.086 $&  $ 1.925 \pm 0.924 $&    $ 44.880 \pm 16.084 $&  DL \\
UT       &  $ 0.838 \pm 0.152 $  &    $ 0.742 \pm 0.161 $&  $ 4.812 \pm 13.604$&    $ 34.634 \pm 12.351 $&  MAS \\
UB1$^*$  &  $ 0.887 \pm 0.030 $  &    $ 0.798 \pm 0.048 $&  $ 1.443 \pm 0.302 $&    $ 55.426 \pm 10.924 $&  DL \\
UOE$^*$  &  $ 0.806 \pm 0.159 $  &    $ 0.697 \pm 0.166 $&  $ 4.197 \pm 7.780 $&    $ 51.922 \pm 17.482 $&  DL \\
\hline \hline
\multirow{3}{*}{Average}
&           $ 0.859 \pm 0.108 $  &    $ 0.763 \pm 0.118 $&  $ 3.259 \pm 9.748 $&    $ 34.382 \pm 12.468 $&  MAS \\
&           $ 0.875 \pm 0.083 $  &    $ 0.784 \pm 0.010 $&  $ 1.840 \pm 2.963 $&    $ 38.510 \pm 17.890 $&  DL \\
&           $ 0.872 \pm 0.087 $  &    $ 0.780 \pm 0.102 $&  $ 2.124 \pm 5.133 $&    $ 37.684 \pm 17.026 $&  ALL \\

\hline
\end{tabular} }
\end{table*}

\begin{table*} [t] \center
    \caption{
    Results of the eleven evaluated algorithms on MRI dataset.
    SD: surface-to-surface distance;
    HD: Hausdorff  Distance;
    DL: deep learning-based method;
    MAS: conventional method based on multi-atlas segmentation.
    Asterisk (*) indicates the results were submitted after the challenge deadline.
    }
\label{tb:table:result_mr}
{\footnotesize
\begin{tabular}{ l| lllll}\hline
Teams & Dice & Jaccard &   SD (mm) & HD (mm) & DL/MAS\\
\hline
UOL      &  $ 0.870 \pm 0.035 $&  $ 0.772 \pm 0.054 $&  $ 1.700 \pm 0.649 $&  \bm{$ 28.535 \pm 13.220 $}&    MAS \\
GUT      &  $ 0.863 \pm 0.043 $&  $ 0.762 \pm 0.064 $&  $ 1.890 \pm 0.781 $&  $ 30.227 \pm 14.046 $&  DL \\
KTH      &  $ 0.855 \pm 0.069 $&  $ 0.753 \pm 0.094 $&  $ 1.963 \pm 1.012 $&  $ 30.201 \pm 13.216 $&  DL \\
UCF      &  $ 0.818 \pm 0.096 $&  $ 0.701 \pm 0.118 $&  $ 3.040 \pm 3.097 $&  $ 40.092 \pm 21.119 $&  DL \\
UT       &  $ 0.817 \pm 0.059 $&  $ 0.695 \pm 0.081 $&  $ 2.420 \pm 0.925 $&  $ 30.938 \pm 12.190 $&  MAS\\
CUHK2    &  $ 0.810 \pm 0.071 $&  $ 0.687 \pm 0.091 $&  $ 2.385 \pm 0.944 $&  $ 33.101 \pm 13.804 $&  DL \\
CUHK1    &  $ 0.783 \pm 0.097 $&  $ 0.653 \pm 0.117 $&  $ 3.233 \pm 1.783 $&  $ 44.837 \pm 15.658 $&  DL \\
SIAT     &  $ 0.674 \pm 0.182 $&  $ 0.532 \pm 0.178 $&  $ 9.776 \pm 6.366 $&  $ 92.889 \pm 18.001 $&  DL \\
UB2$^*$  &  \bm{$ 0.874 \pm 0.039 $}  & \bm{$ 0.778 \pm 0.060 $} & \bm{$ 1.631 \pm 0.580 $}& $ 28.995 \pm 13.030 $&      DL \\
UB1$^*$  &  $ 0.869 \pm 0.058 $&  $ 0.773 \pm 0.079 $&  $ 1.757 \pm 0.814 $&  $ 30.018 \pm 14.156 $&  DL \\
UOE$^*$  &  $ 0.832 \pm 0.081 $&  $ 0.720 \pm 0.105 $&  $ 2.472 \pm 1.892 $&  $ 41.465 \pm 16.758 $&  DL \\
\hline \hline
\multirow{3}{*}{Average}
&          $ 0.844 \pm 0.047  $&  $ 0.734 \pm 0.072 $&  $ 2.060 \pm 0.876 $&  $ 29.737 \pm 12.771 $&  MAS\\
&          $ 0.820 \pm 0.107  $&  $ 0.707 \pm 0.127 $&  $ 3.127 \pm 3.640 $&  $ 41.314 \pm 24.711 $&  DL \\
&          $ 0.824 \pm 0.102  $&  $ 0.711 \pm 0.125 $&  $ 2.933 \pm 3.339 $&  $ 39.209 \pm 23.435 $&  ALL \\
\hline
\end{tabular} }\\
\end{table*}

\subsection{KTH Royal Institute of Technology (KTH)}

\citet{conf/stacom/Wang2017} propose an automatic WHS framework combined CNN with statistical shape priors.
The additional shape information, also called shape context \citep{journal/prl/Mahbod2018}, is used to provide explicit 3D shape knowledge to the CNN. 
The method uses a random forest based landmark detection to detect the ROI.
The statistical shape models are created using the segmentation masks of the 20 training CT images.
The probability map is generated from three 2D U-Nets learned from the multi-view slices of the 3D training images.
To estimate the shape of each subregion of heart, a hierarchical shape prior guided segmentation algorithm \citep{conf/icpr/Wang2014} is then performed on the probability map.
This shape information is represented using volumetric shape models, i.e., signed distance maps of the corresponding shapes.
Finally, the estimated shape information is used as an extra channel, to train a new set of multi-view U-Nets for the final segmentation of whole heart.

\subsection{The Chinese University of Hong Kong, Method No.~1 (CUHK1)}

\citet{conf/stacom/Yang2017a} apply a general and fully automatic framework based on a 3D fully convolutional network (FCN).
The framework is reinforced in the following aspects:
First, an initialization is achieved by inheriting the knowledge from a 3D convolutional networks trained on the large-scale Sports-1M video dataset \citep{conf/piccv/Tran2014}.
Then, the gradient flow is applied by shortening the back-propagation path and employing several auxiliary loss functions on the shallow layers of the network.
This is to tackle the low efficiency and over-fitting issues when directly train the deep 3D FCNs, due to the gradient vanishing problem in shallow layers.
Finally, the Dice similarity coefficient based loss function \citep{conf/3dv/Milletari2016} is included into a multi-class variant to balance the training for all classes.

\begin{figure}[t]\center
 \includegraphics[width=0.48\textwidth]{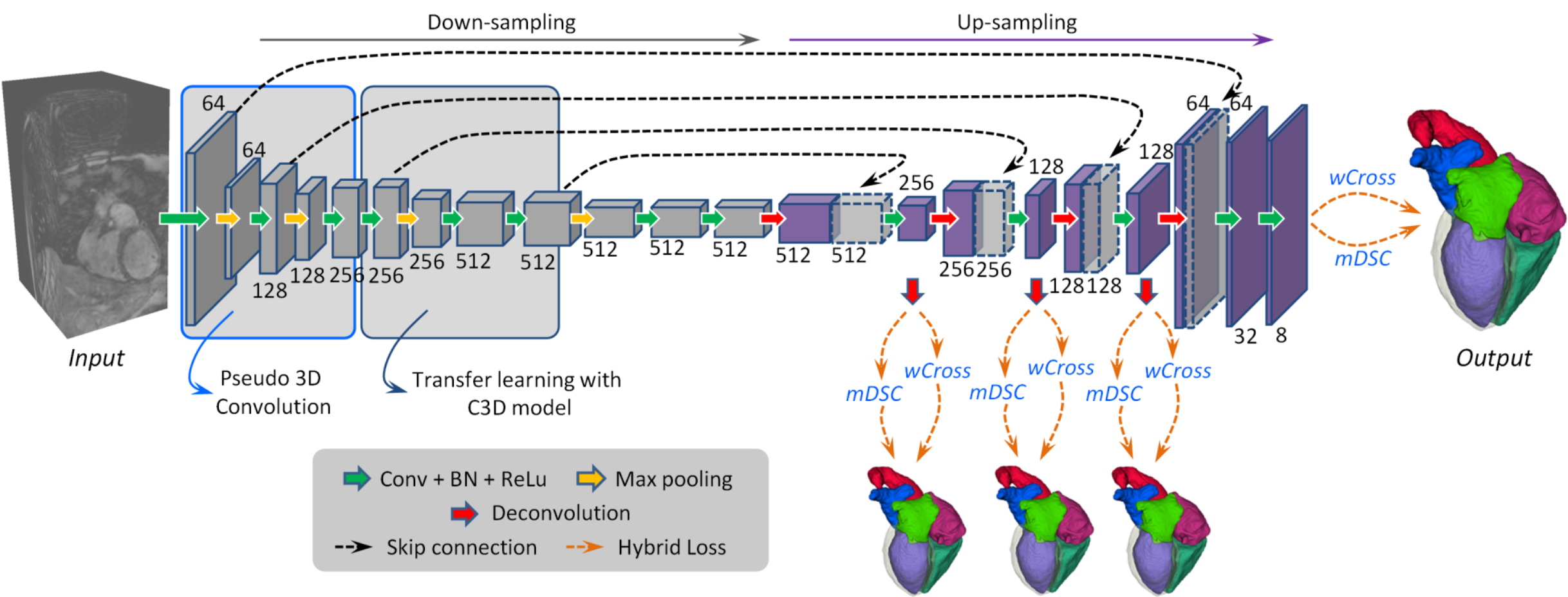}\\[-2ex]
   \caption{A schematic illustration of the method developed by \citet{conf/stacom/Yang2017b}. Digits represent the number of feature volumes in each layer. Volume with dotted line is for concatenation. }
\label{fig:method:CUHK2}\end{figure}

\begin{figure*}[t]\center
 \includegraphics[width=1.01\textwidth]{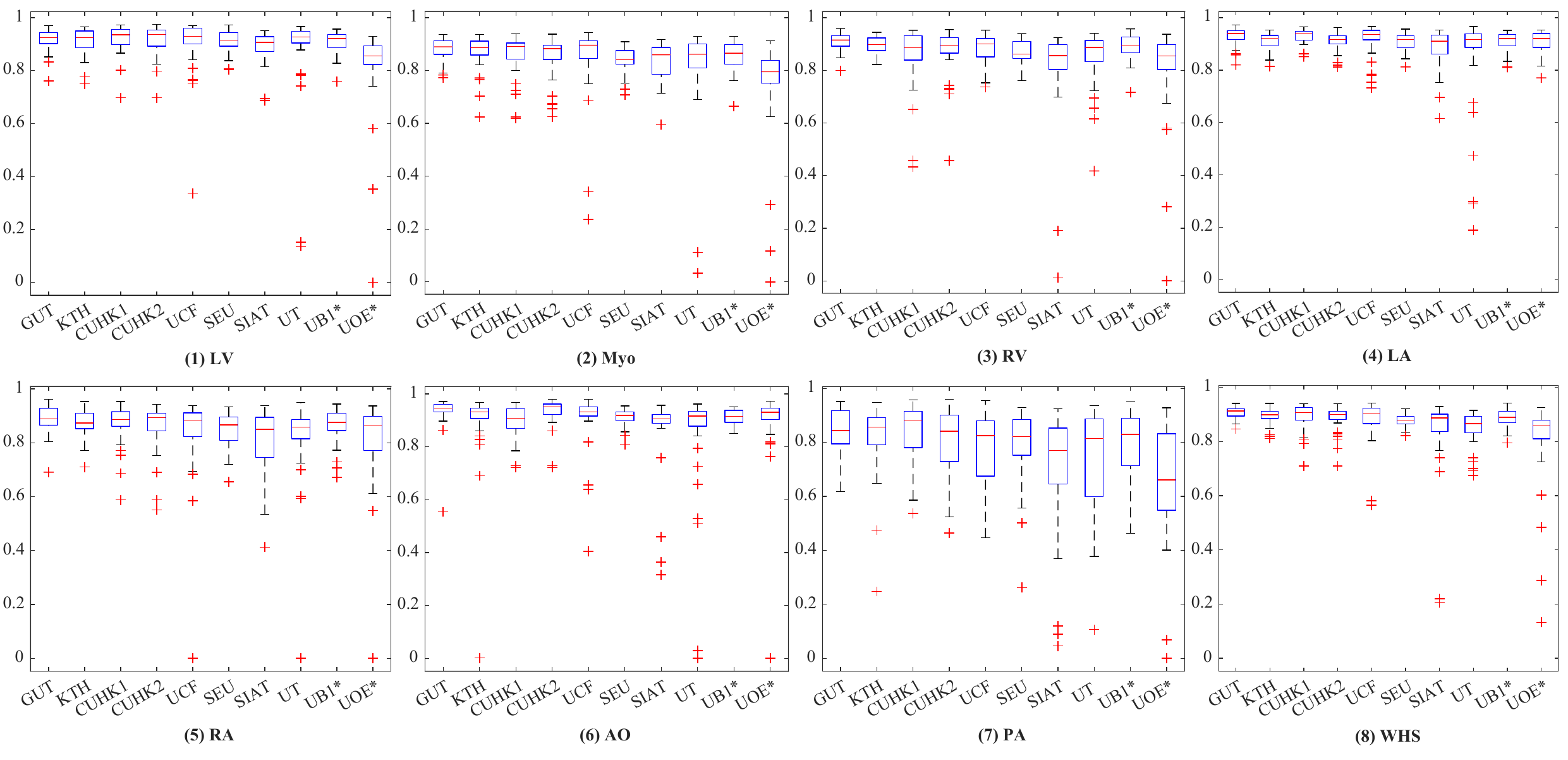}\\[-2ex]
   \caption{Boxplot of Dice scores of the whole heart segmentation on CT dataset by the ten methods.}
\label{fig:ctboxplot}\end{figure*}

\begin{figure*}[t]\center
 \includegraphics[width=1.02\textwidth]{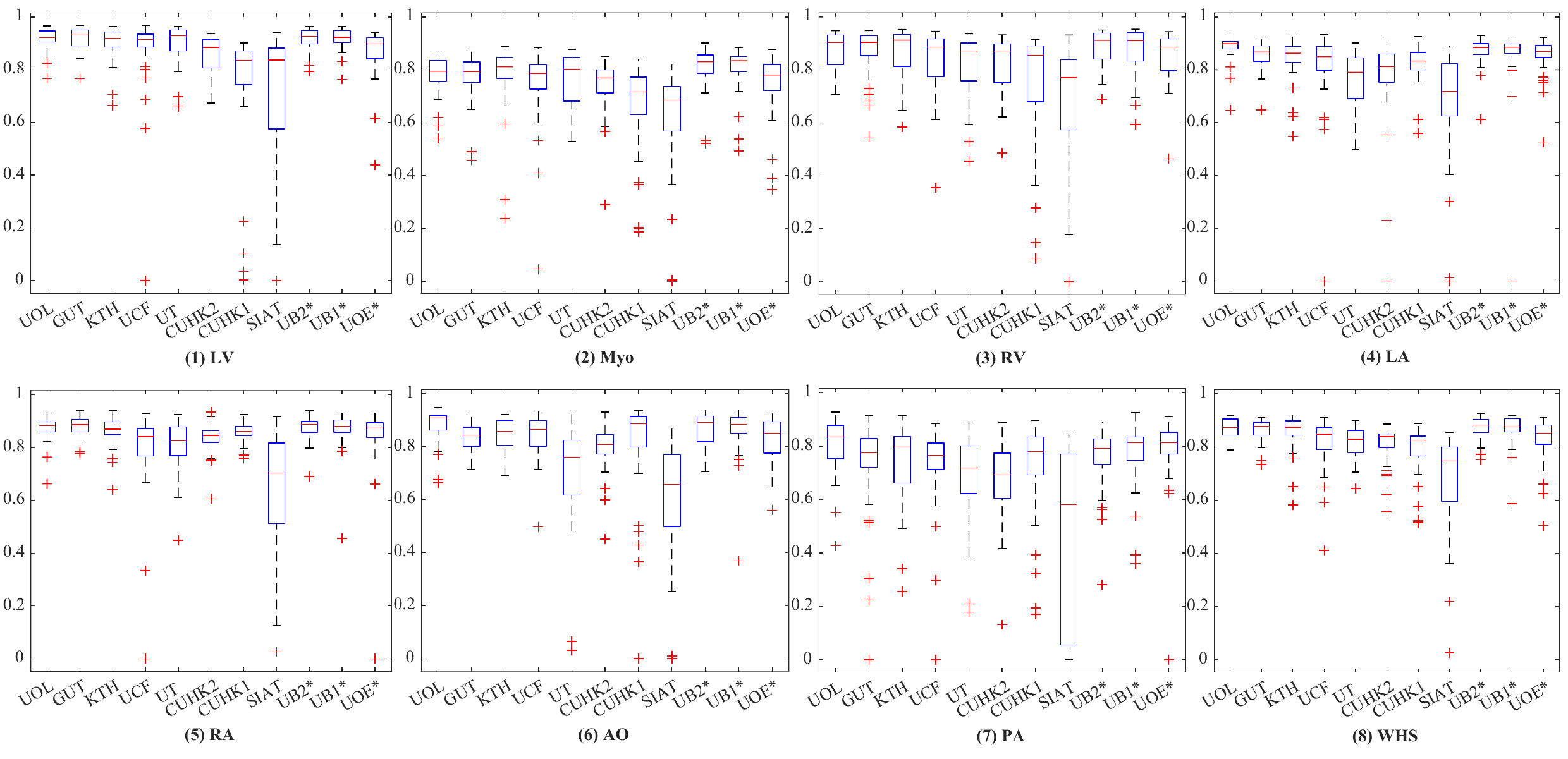}\\[-2ex]
   \caption{Boxplot of Dice scores of the whole heart segmentation on MRI dataset by the eleven methods.}
\label{fig:mrboxplot}\end{figure*}

\subsection{University of Central Florida (UCF)}

\citet{conf/stacom/Mortazi2017} propose a multi-object multi-planar CNN (MO-MP-CNN) method based on an encoder-decoder CNN.
\zxhcolor{lei: Mortazi may just want to add a reference of their miccai paper}{The multiple CNNs \citep{conf/miccai/Mortazi2017} are trained from three different views, i.e., axial, sagittal, and coronal views, in 2D manners.}
An adaptive fusion method is then employed to combine the multiple outputs to refine the delineation.
Furthermore, they apply the connected component analysis (CCA) on the final segmentation, to estimate the reliable (true positive) and unreliable (false positives) regions.
Let $n$ denotes the number of classes in the images and $m$ denotes the number of components in each class, then the CCA could be performed as follows,
\begin{equation}
  \begin{split}
     CCA(S) = \{S_{11},\cdots,S_{nm}|\cup S_{ij}=\textbf{o}\} \&  \\
     \{S_{11},\cdots,S_{nm}|\cap S_{ij}=\phi\},
  \end{split}
\end{equation} where $S$ indicates the segmentation result.
The differences between the reliable and unreliable regions are used to guide the reliability of the segmentation process, namely the higher the difference, the more reliable the segmentation.

\subsection{The Chinese University of Hong Kong, Method No.~2 (CUHK2)}

\citet{conf/stacom/Yang2017b} employ a 3D FCN for an end-to-end dense labeling, as \zxhreffig{fig:method:CUHK2} shows.
The proposed network is coupled with several auxiliary loss functions in a deep supervision mechanism, to tackle the potential gradient vanishing problem and class imbalance in training.
The network learns a spatial-temporal knowledge from a large-scale video dataset, and then transfer to initialize the shallow convolutional layers in the down-sampling path \citep{conf/piccv/Tran2014}.
For the class imbalance issue,
a hybrid loss is proposed \citep{conf/3dv/Milletari2016}, combining two complementary components:
(1) volume-size weighted cross entropy loss (\textit{wCross}) to preserve branchy details such as PA trunks.
(2) multi-class Dice similarity coefficient loss (\textit{mDSC}) to compact anatomy segmentation.
Then, the proposed network can be well trained to simultaneously segment different classes of heart substructures, and generate a segmentation in a dense but detail-preserved format.

\subsection{Southeast University (SEU)}

\citet{conf/stacom/Yang2017c} develop a MAS-based method for WHS of CT images.
The proposed method consists of the following major steps.
Firstly, a ROI detection is performed on atlas images and label images, which are down-sampled and resized to crop and generate a heart mask.
Then, an affine registration is used to globally align the target image with the atlas images, followed by a nonrigid registration to refine alignment of local details.
In addition, an atlas ranking step is applied by using mutual information as the similarity criterion, and those atlases with low similarity are discarded.
A non-rigid registration is further performed by minimizing the dissimilarity within the heart substructures using the adaptive stochastic gradient descent method.
Finally, the propagated labels are fused with different weights according to the similarities between the deformed atlases and the target image.

\subsection{University of Tours (UT)}

\citet{conf/stacom/Galisot2017} propose an incremental and interactive WHS method, combining several local probabilistic atlases based on a topological graph.
The training images are used to construct the probabilistic atlases, for each of the substructures of the heart.
The graph is used to encode the priori knowledge to incrementally extract different ROIs.
The priori knowledge about the shape and intensity distributions of substructures is stored as features to the nodes of the graph.
The spatial relationships between these anatomical structures are also learned and stored as the profile of edges of the graph.
In the case of multi-modality data, multiple graphs are constructed, for example two graphs are built for the CT and MRI images, respectively.
A pixelwise classification method combining hidden Markov random field is developed to integrate the probability map information.
To correct the misclassifications, a post-correction is performed based on the  Adaboost scheme.

\subsection{Shenzhen Institutes of Advanced Technology (SIAT)}

\citet{conf/stacom/Tong2017} develop a deeply-supervised end-to-end 3D U-Net for fully automatic WHS.
The training dataset are artificially augmented by considering each ROI of the heart substructure independently.
To reduce false positives from the surrounding tissues, a 3D U-Net is firstly trained to coarsely detect and segment the whole heart structure.
To take full advantage of multi-modality information so that features of different substructures could be better extracted, the cardiac CT and MRI data are fused.
Both the size and the intensity range of the different modality images are normalized before training the 3D U-Net model.
Finally, the detected ROI is refined to achieve the final WHS, which is performed by a pixel-wise classification fashion using the 3D U-Net.

\subsection{University of Bern, Method No.~1 (UB1*)}

\citet{conf/miccai/Shi2018} design a pixel-wise dilated residual networks, referred to as Bayesian VoxDRN, to segment the whole heart structures from 3D MRI images.
It can be used to generate a semantic segmentation of an arbitrary-sized volume of data after training.
Conventional FCN methods integrate multi-scale contextual information by reducing the spatial resolution via successive pooling and sub-sampling layers,  for semantic segmentation.
By contrast, the proposed method achieves the same goal using dilated convolution kernels, without decreasing the spatial resolution of the network output.
Additionally, residual learning is incorporated as pixel-wise dilated residual modules to alleviate the degrading problem,
and the WHS accuracy can be further improved by avoiding gridding artifacts introduced by the dilation \citep{conf/cvpr/Yu2017}.

\subsection{University of Bern, Method No.~2 (UB2*)}

This method includes a multi-scale pixel-wise fully convolutional Dense-Nets (MSVoxFCDN) for 3D WHS of MRI images, which could directly map a whole volume of data to its volume-wise labels after training.
The multi-scale context and multi-scale deep supervision strategies are adopted,to enhance feature learning.
The deep neural network is an encoder (contracting path)-decoder (expansive path) architecture.
The encoder is focused on feature learning, while the decoder is used to generate the segmentation results.
Skip connection is employed to recover spatial context loss in the down-sampling path.
To further boost feature learning in the contracting path, multi-scale contextual information is incorporated.
Two down-scaled branch classifiers are inserted into the network to alleviate the potential gradient vanishing problem.
Thus, more efficient gradients can be back-propagated from loss function to the shallow layers.

\subsection{University of Edinburgh (UOE*)}

\citet{conf/stacom/Wang2017} develop a two-stage concatenated U-Net framework that simultaneously learns to detect a ROI of the heart and classifies pixels into different substructures without losing the original resolution.
The first U-Net uses a down-sampled 3D volume to produce a coarse prediction of the pixel labels, which is then re-sampled to the original resolution.
The architecture of the second U-Net is inspired by the SRCNN \citep{journal/pami/Dong2016} with skipping connections and recursive units \citep{conf/cvpr/Kim2015}.
It inputs a two-channel 4D volume, consisting of the output of the first U-Net and the original data.
In the test phase, a dynamic-tile layer is introduced between the two U-Nets to crop a ROI from both the input and output volume of the first U-Net.
This layer is removed when performing end-to-end training to simplify the implementation.
Unlike the other U-Net based architecture, the proposed method can directly perform prediction on the images with their original resolution, thanks to the SRCNN-like network architecture.

\section{{Results}}\label{result}

\Zxhreftb{tb:table:result_ct} and \Zxhreftb{tb:table:result_mr} present the quantitative results of the evaluated algorithms on CT and MRI dataset, respectively.

For the CT data, the results are generally promising, and the best Dice score ($0.91\pm0.09$) was achieved by GUT, which is a DL-based algorithm with anatomical label configurations.
The DL-based methods generally obtained better accuracies than the MAS-based approaches in terms of Jaccard, Dice, and SD metrics, though this conclusion was not applied when the HD metric is used.
Particularly, one can find that the mean of HD from the two MAS methods was not worse than that of the other eight DL-based approaches.

For MRI data, the best Dice score of the WHS ($0.87\pm0.04$) was obtained by UB2$^*$, which is a DL-based method and a delayed submission; and the best HD ($ 28.535 \pm 13.220 $ mm) was achieved by UOL, a MAS-based algorithm.
Here, the average accuracy of MAS (two teams) was better than that of the DL-based segmentation (nine teams) in all evaluation metrics.
However, the performance across different DL methods could vary a lot, similar to the results from the CT experiment.
For example, the top four DL methods, i.e., GUT, KTH, UB1$^*$ and UB2$^*$, obtained comparable accuracy to that of UOL,
but the other DL approaches could generate much poorer results.


\zxhreffig{fig:ctboxplot} shows the boxplots of the evaluated algorithms on CT data. One can see that they achieved relatively accurate segmentation for all substructures of the heart, except for the PA whose variability in terms of shape and appearance is notably greater.
For GUT, KTH, CUHK1, UB1$^*$, and CUHK2, the delineation of PA is reasonably good with the mean Dice score larger than 0.8.
\zxhreffig{fig:mrboxplot} presents the boxplots on the MRI data.
The five methods, i.e., UB2$^*$, UOL, UB1$^*$, GUT, and KTH, all demonstrate good Dice scores on the segmentation of four chambers and LV myocardium.
Similar to the conclusion drawn  from \Zxhreftb{tb:table:result_ct} and \Zxhreftb{tb:table:result_mr}, the segmentation on the CT images is generally better than that on the MRI data as indicated by the quantitative evaluation metrics.

\begin{figure*}[t]\center
 \includegraphics[width=1\textwidth]{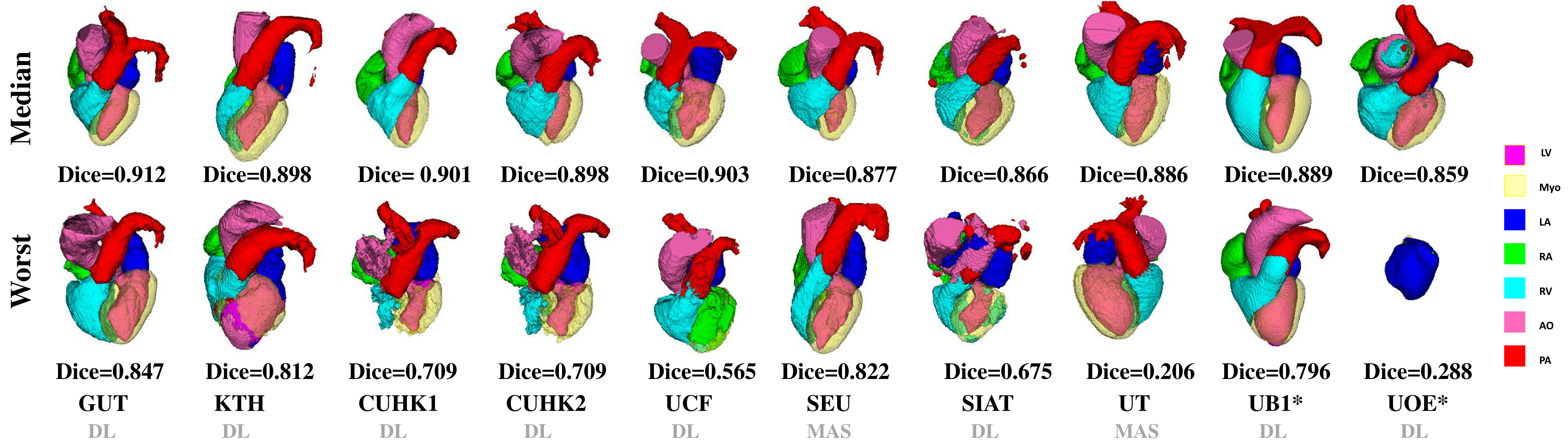}
   \caption{3D visualization of the WHS results of the median and worse cases in the CT test dataset by the ten evaluated methods.
   The color bar indicates the correspondence of substructures.
   Note that the colors of Myo and LV in 3D visualization do not look exactly the same as the keys in the color bar, due to the 50\% transparency setting for Myo rendering and the addition effect from two colors (LV and 50\% Myo) for LV rendering, respectively.
   }
\label{fig:ct_3dvisual}\end{figure*}

\begin{figure*}[t]\center
 \includegraphics[width=1\textwidth]{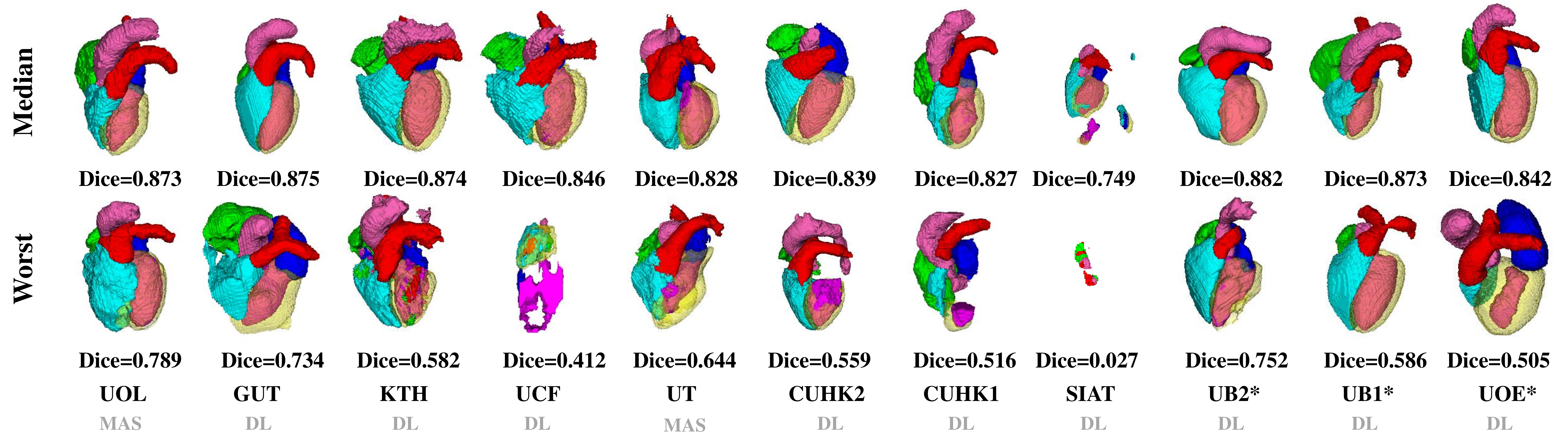}
   \caption{3D visualization of the WHS results of the median and worse cases in the MRI test dataset by the eleven evaluated methods.
   }
\label{fig:mr_3dvisual}\end{figure*}

\zxhreffig{fig:ct_3dvisual} shows the 3D visualization of the cases with the median and worst WHS Dice scores by the evaluated methods on the CT data.
Most of the median cases look reasonablely good, though some contain patchy noise; and the worst cases require significant improvements.
Specifically, 
UOE$^*$ median case contains significant amount of misclassification in AO,
and parts of the LV are labeled as LA in the UOE$^*$ and SIAT median cases.
In the worst cases,
the CUHK1 and CUHK2 results do not have a complete shape of the RV;
KTH and SIAT contain a large amount of misclassification, particularly in myocardium;
UCF mistakes the RA as LV;
UOE$^*$ only segments the LA, and UT generates a result with wrong orientation.

\zxhreffig{fig:mr_3dvisual} visualizes the median and worst results on MRI WHS.
Compared with the CT results, even the median cases of MRI cases are poor.
For example, the SIAT method could perform well on most of the CT cases, but failed to generate acceptable results for most of the MRI images, including the median case presented in the figure.
The worst cases of UOE$^*$, CUHK2 and UB1 miss at least one substructure,
and UCF and SIAT results do not contain any complete substructure of the whole heart.
In conclusion, the CT segmentation results look better than the MRI results, which is consistent with the quantitative results.
Also, one can conclude from \zxhreffig{fig:ct_3dvisual}  and \zxhreffig{fig:mr_3dvisual} that the resulting shape from the MAS-based methods looks more realistic, compared to the DL-based algorithms, even though the segmentation could sometimes be very poor or even a failure, such as the worst MRI case by UOL and the worst CT case by UT.

\begin{table*}[!tb]\center
   \caption{Summary of the advantages and limitations of the 12 benchmarked methods.} \label{tb:summary}
\begin{tabular}{@{}c@{}}
 \includegraphics[width=\textwidth]{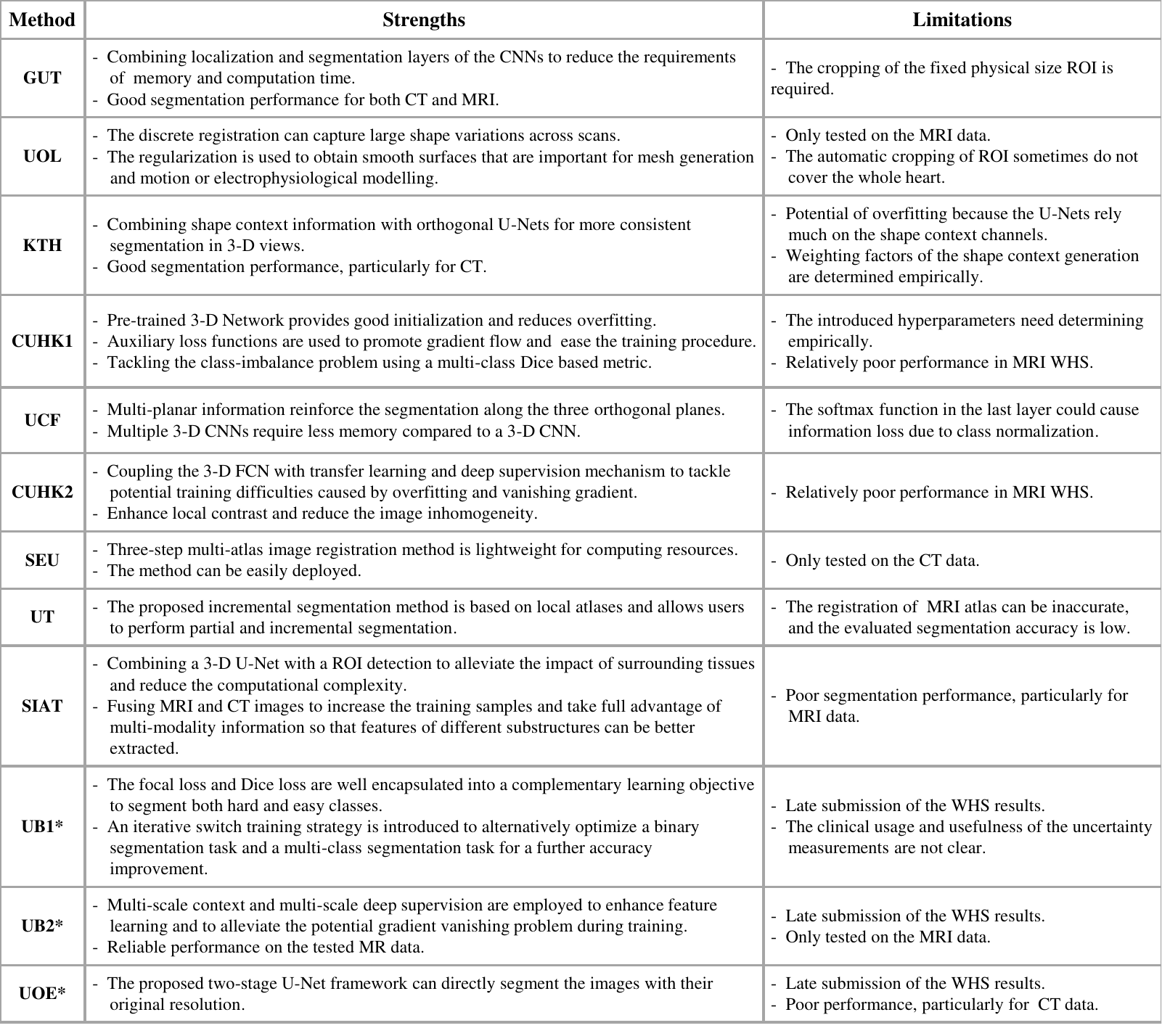}
\end{tabular}\end{table*}

\section{Discussion}\label{discussion}


\subsection{Overall performance of the evaluated algorithms} \label{overall perf}

The mean Dice scores of the evaluated methods for MM-WHS are respectively
$0.872\pm0.087$ (CT) and $0.824\pm0.102$ (MRI), and the best average Dices from one team are respectively $0.908\pm0.086$ (CT by GUT) and $0.874\pm0.039$ (MRI by UB2$^*$).
\zxhreftb{tb:table:result_ct} and \zxhreftb{tb:table:result_mr} provide the average numbers of the other evaluation metrics, for the different methodological categories and different imaging modalities.
In general, the benchmarked algorithms obtain better WHS accuracies for CT than for MRI, using the four metrics.
In addition, the mean Dice scores of MAS-based methods are $0.859\pm0.108$ (CT) and $0.844\pm0.047$ (MRI), and those of DL-based methods are $0.875\pm0.083$ (CT) and $0.820\pm0.107$ (MRI).
DL-based WHS methods obtain better mean accuracies,
 but the MAS-based approaches tend to generate results with more realistic heart shapes.

Furthermore, the segmentation accuracies reported for the four chambers are generally good, but the segmentation of the other substructures demonstrates more challenges.
For example, one can see from \zxhreffig{fig:ctboxplot} and \zxhreffig{fig:mrboxplot} that
 in CT WHS the PA segmentation is much poorer compared to other substructures;
 in MRI WHS, the segmentation of myocardium, AO and PA appears to be more difficult.
One reason could be that these regions have much larger variation in terms of shapes and image appearance across different scans.
Particularly, the diverse pathologies can result in heterogeneous intensity of the myocardium and blood fluctuations to the great vessels.
The other reason could be the large variation of manual delineation of boundaries for these regions, which results in more ambiguity for the training of learning-based algorithms and the generation of the gold standard.

\subsection{MAS versus DL-based segmentation}


As \zxhreftb{tb:table:result_ct} and \zxhreftb{tb:table:result_mr} summarize, 9 out of the 11 benchmarked CT WHS methods and 8 out of the 10 MRI WHS algorithms are based on deep neural networks.
In general, the DL-based approaches can obtain good scores when the models have been successfully trained.
However, tuning the parameters for a network to obtain the optimal performance can be difficult, as several DL-based methods reported poor results.
This is also evident from \zxhreffig{fig:ctboxplot} and \zxhreffig{fig:mrboxplot} where some of the DL methods have very large interquartile ranges and outliers,
 and from the 3D visualization results presented in  \zxhreffig{fig:ct_3dvisual} and \zxhreffig{fig:mr_3dvisual}.
In several cases, the shape of the heart from the segmentation results can be totally unrealistic, such as
the worst CT case of UOE$^*$, median and worst MRI cases of SIAT, worst MRI cases of CUHK1 and UCF.

In general, the conventional methods, mainly based on MAS framework, can generate results with more realistic shapes, though their mean accuracies can be less compared to the well trained DL models.
Particularly, in MRI WHS the MAS-based methods obtained better mean accuracies than the DL-based approaches, though only two MAS methods were submitted for comparisons.
Notice that the WHS of MRI is generally considered more challenging compared to that of CT.
Since the DL-based approaches performed much better in the CT WHS, one can expect the performance of MR WHS could be significantly improved by resorting to new DL technologies in the future.


\subsection{CT WHS versus MRI WHS}
The MRI WHS is generally more arduous than the CT WHS, which is confirmed by the results presented in this work.
The mean generalized Dice score of CT WHS is evidently better than that of MRI WHS averaged from the benchmarked algorithms, namely $0.872\pm0.087$ (CT) versus $0.824\pm0.102$ (MRI).
One can further confirm this by comparing the results for these two tasks in \zxhreftb{tb:table:result_ct} and \zxhreftb{tb:table:result_mr}, as nine methods have been evaluated on both the CT and MRI test data, and the same algorithms generally obtain better accuracies for CT data.
Similar conclusion can be also drawn for the individual substructures as well as for the whole heart, when one compares the boxplots of segmentation Dice scores between \zxhreffig{fig:ctboxplot} and \zxhreffig{fig:mrboxplot}.




\subsection{Progress and challenges}

The MM-WHS challenge provides an open access dataset and ongoing evaluation framework for researchers,
who can make full use of the open source data and evaluation platform to develop and compare their algorithms.
Both the conventional methods and the new DL-based algorithms have made great progress shown in this paper.
It is worth mentioning that the DL models with best performance have demonstrated potential of generating accurate and reliable WHS results, such as the methods from GUT, UB1$^*$ and UB2$^*$, though they were trained using 40 training images (20 CT and 20 MRI).
Nevertheless, there are limitations, particularly from the methodological point of view.
\Zxhreftb{tb:summary} summarizes the advantages and potential limitations of the benchmarked works.

WHS of MRI is more arduous. The average performance of the MRI WHS methods is not as good as that of the CT methods, concluded from the submissions.
The challenges could mainly come from the low image quality and inconsistent appearance of the images, as well as the large shape variation of the heart which CT WHS also suffers from.
Enlarging the size of training data is a commonly pursued means for improving the learning-based segmentation algorithms.
However, availability of whole heart training images can be as challenging as the task itself.
One potential solution is to use artificial training data, such as by means of data augmentation or image synthesis using generative adversarial networks \citep{conf/anips/2014}.
Alternately, shape constraints can be incorporated into the training and prediction framework, which is particularly useful for the DL-based methods to avoid generating results of unrealistic shapes.

\section{Conclusion}\label{conclusion}

Knowledge of the detailed anatomy of the heart structure is clinically important as it is closely related to cardiac function and patient symptoms.
Manual WHS is labor-intensive and also suffers from poor reproducibility.
A fully automated multi-modality WHS is therefore highly in demand.
However, achieving this goal is still challenging, mainly
due to the low quality of whole heart images, complex structure of the heart and large variation of the shape.
This manuscript describes the MM-WHS challenge which provides 120 clinical MRI/ CT images,
 elaborates on the methodologies of twelve evaluated methods,
 and analyzes their evaluated results.


The challenge provides the same training data and test dataset for all the submitted methods.
Note that these data are also open to researchers in future.
The evaluation has been performed by the organizers, blind to the participants for a fair comparison.
The results show that WHS of CT has been more successful than that of MRI from the twelve submissions.
For segmentation of the substructures, the four chambers generally are easy to segment from the submitted results.
By contrast, the great vessels, including aorta and pulmonary artery, still need more efforts to achieve good results.
For different methodologies, the DL-based methods could achieve high accuracy for the cases they succeed.
They could also generate poor results with unrealistic shape, namely the performance can vary a lot.
The conventional atlas-based approaches, either using segmentation propagation or probabilistic atlases, however generally perform stably, though they are not as widely used as the DL technology now.
The hybrid methods, combining deep learning with prior information from either the multi-modality atlas or shape information of the heart substructures, should have potential and be worthy of future exploration.

\section*{Authors contributions}
XZ initialized the challenge event, provided the 60 CT images,
41 MRI images (with KR and SO) of the 60 MRI images,
and the manual segmentations of all the 120 images.
GY, RM, JK, and DF provided the other 19 MRI images.
XZ, GY and LL organized the challenge event,
 and LL evaluated all the submitted segmentation results.
GY generated the first draft, based on which XZ and LL restructured and rewrote the manuscript.
CP, DS, MU, MPH, JO, CW, OS, CB, XY, PAH, AM, UB, JB, GYu, CS, GG, JYR, TB, QT, WS, and XL were the participates of the MM-WHS challenge and contributed equally.
GZ, ZS, GZ, CW, TM and DN submitted their results after the deadline of the challenge.
All of the participants provided their results for evaluation and the description of their algorithms.
All authors have read and approved the publication of this work.

\section*{Acknowledgement}This work was funded in part by the Chinese NSFC research fund, the Science and Technology Commission of Shanghai Municipality (17JC1401600) and the British Heart Foundation Project Grant (PG/16/78/32402).

\bibliographystyle{model2-names}
\bibliography{AllBibliography_MMWHS_new}

\end{document}